\documentclass[10pt,twocolumn,letterpaper]{article}

\usepackage{cvpr}
\usepackage{times}
\usepackage{epsfig}
\usepackage{graphicx}
\usepackage{amsmath}
\usepackage{amssymb}

\usepackage{xcolor} 
\usepackage{multirow}
\usepackage{enumerate}
\usepackage{url}
\usepackage{graphicx}
\usepackage{caption}
\usepackage{subcaption}
\usepackage{tabularx}
\usepackage{enumitem}
\usepackage{array}
\usepackage{float}

\usepackage[pagebackref=true,breaklinks=true,letterpaper=true,colorlinks,bookmarks=false]{hyperref}

\cvprfinalcopy 

\def\cvprPaperID{5} 

\ifcvprfinal\fi
\begin{document}

\title{Deep Anomaly Detection for Generalized Face Anti-Spoofing}

\author{Daniel P\'erez-Cabo\\
Gradiant - UVigo, Spain\\
{\tt\small dpcabo@gradiant.org}
\and
David Jim\'enez-Cabello\\
Gradiant, Spain\\
{\tt\small djcabello@gradiant.org}
\and
Artur Costa-Pazo\\
Gradiant, Spain\\
{\tt\small acosta@gradiant.org}
\and
Roberto J. L\'opez-Sastre\\
University of Alcal\'a, Spain\\
{\tt\small robertoj.lopez@uah.es}
}

\maketitle
\thispagestyle{empty}

\begin{abstract}
  Face recognition has achieved unprecedented results, surpassing human capabilities in certain scenarios. 
However, these automatic solutions are not ready for production because they can be easily fooled by simple identity impersonation attacks. And although much effort has been devoted to develop face anti-spoofing models, their generalization capacity still remains a challenge in real scenarios. In this paper, we introduce a novel approach that reformulates the Generalized Presentation Attack Detection (GPAD) problem from an anomaly detection perspective. Technically, a deep metric learning model is proposed, where a triplet focal loss is used as a regularization for a novel loss coined ``metric-softmax'', which is in charge of guiding the learning process towards more discriminative feature representations in an embedding space. Finally, we demonstrate the benefits of our deep anomaly detection architecture, by introducing a few-shot a posteriori probability estimation that does not need any classifier to be trained on the learned features. 
We conduct extensive experiments using the GRAD-GPAD framework that provides the largest aggregated dataset for face GPAD. Results confirm that our approach is able to outperform all the state-of-the-art methods by a considerable margin.
\end{abstract}


\section{Introduction}
\label{sec:introduction}
Whether we like it or not, we are in the era of face recognition automatic systems. These solutions are now beginning to be used intensively in: border controls, on-boarding processes, accesses to events, automatic login, or to unlock our mobile devices. As an example of this last technology, we have the \emph{Intelligent Scan}\footnote{\scriptsize{\url{https://www.samsung.com/my/support/mobile-devices/what-is-intelligent-scan-and-how-to-use-it/}}} that comes with Samsung mobiles, or the \emph{FaceID}\footnote{\scriptsize{\url{https://www.apple.com/lae/iphone-xs/face-id/}}} for iPhones. All these systems are highly valued by consumers because of their usability and its non-intrusive nature. However, there remains one major challenge for all of them, Presentation Attacks (PA).

\begin{figure}[ht]
\begin{center}
\includegraphics[width=0.9\linewidth]{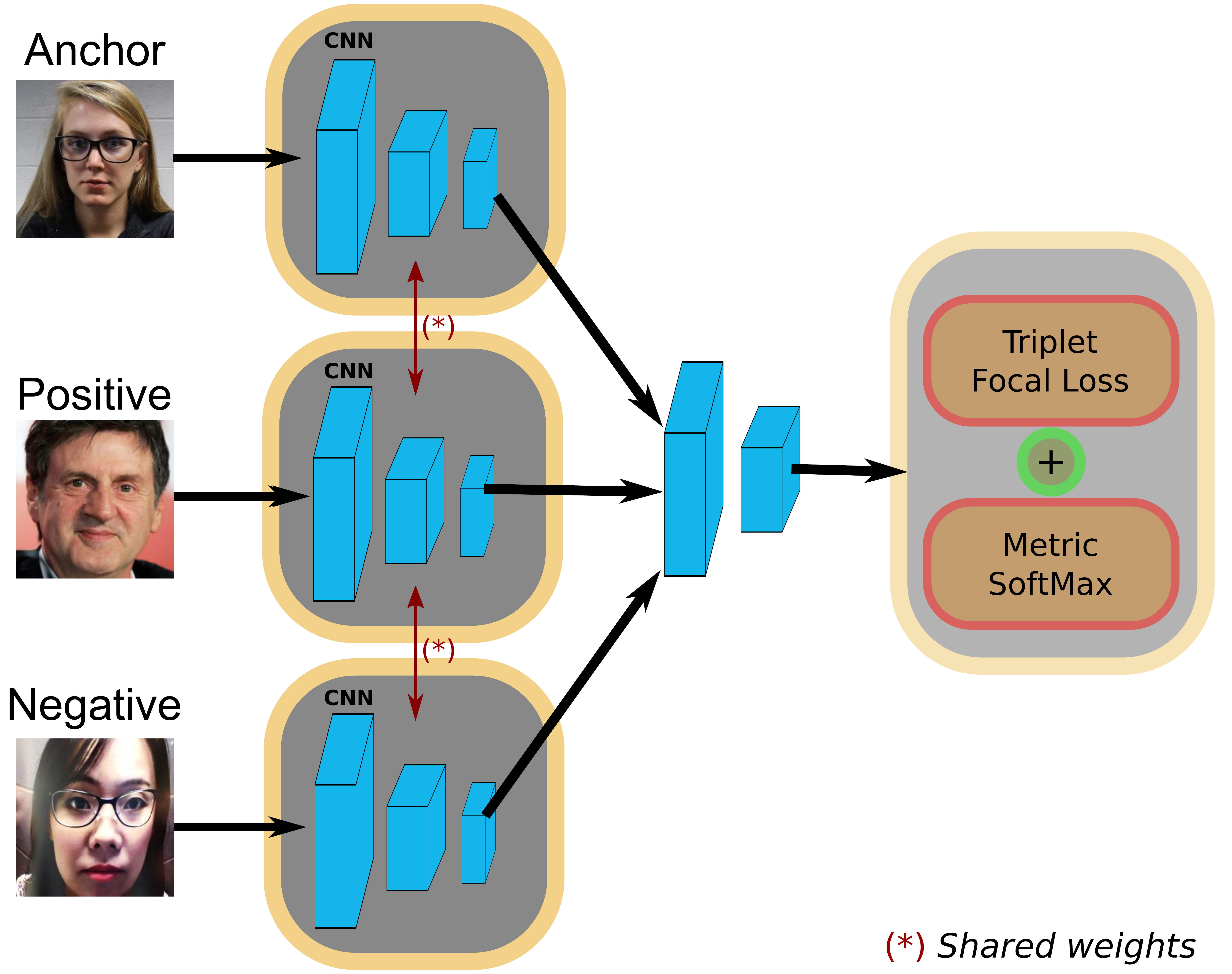}
\end{center}
\caption{We propose a deep metric learning approach, using a set of Siamese CNNs, in conjunction with the combination of a triplet focal loss and a novel ``metric softmax'' loss. The latter accumulates the probability distribution of each pair within the triplet. Our aim is to learn a feature representation that allows us to detect impostor samples as anomalies.}
\label{fig:graphical_abstract}
\end{figure}

These commercial systems rely on specialized hardware such as 3D/IR/thermal cameras entailing a far easier option to detect presentation attacks. Besides, this situation restricts the use case to a few specialized devices, incrementing costs dramatically. For the sake of accessibility and costs, we focus on the ubiquitous 2D-camera case, available in almost all mobile devices and easy to acquire and integrate on different checkpoints.

Although face recognition technologies achieve accuracy ratios above human performance in certain scenarios, consumers should be aware that they also introduce two new challenges that compromise their security: the Presentation Attack Detection (PAD) and the generalization capability of these solutions. With respect to the former, for example, a face recognition system with an outstanding 99.9\% of accuracy fails simply by presenting a page with your face printed on it. These presentation attacks stand as a major threat for identity impersonation where illegitimate users attempt to gain access to a system using different strategies, \eg video replay, make-up. Note that it is really easy to obtain audiovisual material from almost every potential user (\eg Facebook photos, videos on YouTube, \etc), which allows the creation of tools to perform these PAs. 

But the generalization problem is also relevant. In a nutshell, the scientific community has failed to provide an efficient method to detect identity impersonation based on face biometrics that is valid for real-world applications. Normally, the state-of-the-art models suffers a severe drop of performance in realistic scenarios, because they exhibit a sort of overfitting behaviour maximizing the results for just the dataset they have been trained on.

In this paper we explicitly address these two challenges. First, we introduce a deep metric learning based approach to deal with the PAD problem. As it is shown in Fig.~\ref{fig:graphical_abstract}, our solution is trained to learn a feature representation that guarantees a reasonable separability between genuine and impostor samples. Then, the generalization problem is tackled from an anomaly detection approach, where we expect to detect the attacks as if they were out-of-distributions samples that naturally exhibit a higher distance in the embedding space with respect to the real samples in the dataset.

The generalization capability of our solution and its state-of-the-art competitors is thoroughly evaluated using the recent GRAD-GPAD framework~\cite{grad_gpad_gradiant_2019}. We use the aggregated dataset provided in GRAD-GPAD, which comprises more than 10 different datasets for face anti-spoofing. This aspect results fundamental, because it allows us to deploy extensive inter-dataset experiments, to address the Generalized Presentation Attack Detection problem. 

As a summary, in this paper we make the following contributions:
\begin{enumerate}
 \item We introduce a novel anomaly detection strategy based on deep metric learning for face anti-spoofing using just still images.
 \item Our model leverages the use a triplet focal loss as a regularizer of a novel ``metric softmax'' loss, to ensure that the learned features allow for a reasonable separability between real and attacks samples in an embedding space.
 \item A thorough experimental evaluation on GRAD-GPAD shows that our anomaly detection based approach outperforms the state-of-the-art models.
 \item Finally, we propose a novel few-shot a posteriori probability estimation that avoids the necessity of training any classifier for decision making.
\end{enumerate}

The remainder of this paper is organized as follows. Section~\ref{sec:related_work} reviews the main progress and challenges on the problem of generalization for anti-spoofing systems. We introduce our anomaly detection deep model in Section~\ref{sec:approach}. Sections~\ref{sec:experiments} and~\ref{sec:conclusions} provide the experimental evaluation and the conclusions, respectively.

\section{Related Work}
\label{sec:related_work}

Face-PAD approaches can be categorized regarding the following standpoints: i) from the required user interaction as \emph{active}~\cite{kollreider2007real} or \emph{passive}~\cite{db_msu_mfsd_2015, li2018learning} methods; ii) from the hardware used for data acquisition as \emph{rgb-only}~\cite{db_msu_mfsd_2015, iqa_galbally_2015, anomaly_idiap_2018}, \emph{rgb-infrared-depth}~\cite{db_csmad_2018, db_casia_surf_2018} or \emph{additional sensors}~\cite{light_field_based_sepas_2018} approaches; iii) from the input data type as~\emph{video-based}~\cite{motion_anjos_2011, db_uvad_2015} or \emph{single-frame}~\cite{db_msu_mfsd_2015, db_msu_mfsd_2015} approaches; iv) and, finally, depending on the feature type, from classical hand-crafted features~\cite{db_msu_mfsd_2015, boulkenafet2018generalization} to the newer ones based on automatic learned deep features~\cite{li2018learning, db_siw_2018}. These deep models are precisely the responsible for a considerable increase in accuracy for face-PAD, defining the new state of the art.

However, recent studies reveal that the current approaches are not able to correctly generalize \cite{db_rose_youtu_2018} using fair comparisons. Actually, the main difficulty for the inclusion 
of anti-spoofing systems in realistic environments is the Generalized Presentation Attack Detection (GPAD) problem. Some works~\cite{challenges_face_pad_gradiant_2019, anomaly_idiap_2018, grad_gpad_gradiant_2019} propose new evaluation protocols, datasets and methods to address the GPAD.

Overall, generalization has been addressed from different perspectives: i) applying \emph{domain adaptation} techniques~\cite{db_rose_youtu_2018}; ii) learning \emph{generalized deep features}~\cite{db_rose_youtu_2018, li2018learning}; or even iii) using generative models~\cite{db_siw_2018}. All these methods are able to slightly mitigate the drop of performance when testing on new unseen scenarios, but they are still far from being suitable for real scenarios.

Traditional methods for face anti-spoofing use a two-class classifier to distinguish between real samples and attacks. Recently, some works suggest that formulating the problem of anti-spoofing as an anomaly detection approach could improve their generalization capacity~\cite{anomaly_surrey_2017, anomaly_idiap_2018}. In~\cite{anomaly_surrey_2017}, the authors assume that real-accesses share the same nature, in contrast to spoofing attempts that can be very diverse and unpredictable. They present a study to determine the influence of using only genuine data for training and compare it with traditional two-class classifiers. From the experimental results the paper concludes that: i) anomaly detection based systems are comparable to two classes based systems; and ii) neither of the two approaches perform well enough in the evaluated datasets (CASIA-FASD~\cite{db_casia_fasd_2012}, Replay-Attack~\cite{db_replay_attack_2012} and MSU-MFSD~\cite{db_msu_mfsd_2015}). On the other hand, the authors of~\cite{anomaly_idiap_2018} propose a more challenging experiment based on an aggregated dataset that comprises Replay-Attack, Replay-Mobile~\cite{db_replay_mobile_2016} and MSU-MFSD. They propose a GMM-based anomaly classifier which outperforms the best solutions reported in~\cite{anomaly_surrey_2017}.

In this paper, we reformulate the anomaly detection scheme using a deep metric learning model for face-PAD that highly reduces the problem of generalization. Experiments are performed over the largest aggregated publicly available dataset, the GRAD-GPAD framework~\cite{grad_gpad_gradiant_2019}. This framework allows us to reinforce the assumption that real access data shares the same nature, provided that the number of identities is large and the capture conditions and devices are diverse enough; that is, the genuine class is well represented by data. Additionally, the highly representative embeddings obtained using the proposed metric learning approach permits outperforming prior works, distinguishing genuine amongst an open-set class of attacks in the most challenging dataset so far.

\section{Deep Anomaly Detection for Face GPAD}
\label{sec:approach}

\subsection{Review on Metric Learning}
Many works rely on a softmax loss function to separate samples from different classes in deep learning models. However, class compactness is not explicitly considered and samples from different classes might easily overlap in the feature space. Instead, metric learning based losses are designed to address these issues, by promoting inter-class separability and reducing intra-class variance. Note that several metric learning approached have been applied to multiple tasks such as face recognition~\cite{parkhi2015deep}, object retrieval~\cite{he2018triplet} or person re-identification~\cite{Zhang18_tfl}, obtaining outstanding generalization performance. In this section we introduce the mathematical notation and our formulation for the problem of deep anomaly detection for face GPAD, from a metric learning perspective.

Let $f_\theta(x_i)$ be the feature vector in the embedding space of a data point $x_i \in \mathbb{R}^N$, where the mapping function $f_\theta: \mathbb{R}^N \rightarrow \mathbb{R}^D$ is a differentiable deep neural network of parameters $\theta$, and let $D_{i,j}$ be the squared l2-norm between two feature vectors defined by $D_{i, j} = \|f_\theta(x_i) - f_\theta(x_j) \|^2_2$. Usually, $f_\theta(x_i)$ is normalized to have unit length for training stability. In a deep metric learning based approach, the objective is to learn a deep model that generates a feature representation $f_\theta(x_i)$ to guarantee that samples from the same class are closer in the embedding space, than samples from different categories. For doing so, different loss functions can be found in the literature.

For instance, the center loss proposed in~\cite{wen2016discriminative} concentrates samples around their class centers in the embedding space (see Eq.~\ref{eq:center_loss}). It is used in conjunction with the softmax loss to increase intra-class compactness, however the latter does not guarantee a correct inter-class separation.

\begin{equation}
  \mathcal{L}_{c}(\theta) = \displaystyle{\frac{1}{2} \sum_{i=1}^b  \| f_\theta(x_i)- c_{y^i}\|^2_2},
\label{eq:center_loss}
\end{equation}
\noindent where $b$ is the number of input tuples in a batch and $c_{y^i}$ is the class center corresponding to the ground truth label $y^i$ of sample $x_i$.

The contrastive loss~\cite{chopra2005learning} (see Eq.~\ref{eq:contrastive_loss}) forces all images belonging to the same class to be close, while samples from different classes should be separated by a margin $m$. It uses tuples of two images as different image pairs \{$p$, $q$\}: i) \emph{positive}, if both belong to the same class and ii) \emph{negative}, otherwise. However, one needs to fix a constant margin $m$ for the negative pairs, separating all negative examples by the same margin regardless their visual appearance:
\begin{equation}
  \small{\mathcal{L}_{ct}(\theta) = \sum_{i=1}^b  y_{p_i, q_i} D_{p_i, q_i} + (1-y_{p_i, q_i}) \max{\left(0, m-D_{p_i, q_i}\right)}^2},
\label{eq:contrastive_loss}
\end{equation}
\noindent where $y_{p_i, q_i}=1$ for the positive pair and $y_{p_i, q_i}=0$ for the negative.

Following the same idea, the authors of the triplet loss~\cite{TripletLoss_Weinberger} extend the contrastive loss to consider positive and negative pairs simultaneously by using a tuple of three images: i) anchor, ii) positive and iii) negative. The goal of the triplet loss in Eq.~\ref{eq:triplet_loss} is to reduce the intra-class variance defined by the anchor-positive pair, while simultaneously increase the inter-class separation by maximizing the euclidean distance between the anchor-negative pair. Despite avoiding a constant margin for the negative pair and obtaining highly discriminative features, it suffers from the complexity of the triplet selection procedure. Nevertheless, it has been successfully addressed in many recent approaches~\cite{yu2018correcting,ge2018deep,smirnov2018hard}.
 \begin{align}
  \begin{split}\label{eq:triplet_loss}
    \mathcal{L}_{\text{t}}(\theta) = \displaystyle{\sum_{i=1}^b\max{\left(0, D_{a_{i},p_{i}} - D_{a_{i},n_{i}})+m\right)}},
  \end{split}
\end{align}
where $\{a_{i},p_{i},n_{i}\}$ sub-indexes are the anchor, the positive and the negative samples for each triplet within the batch, respectively.

Prior works successfully applied the triplet loss (or any of its variants) using a large number of classes, \eg face recognition models use thousands of identities, for instance in VGG2 Face data set~\cite{VGG2Cao18} there are more than 9000 different identities. Such a diversity of classes encourages embeddings to generalize when the number of samples is large enough. In this paper, we show that a triplet loss based model, following an anomaly detection perspective, can actually outperform existing methods for face GPAD.

\subsection{Triplet Focal Loss for Anomaly Detection for Face GPAD}
We address the face GPAD problem from a metric learning approach with a Triplet focal loss. Technically, we propose to use a modified version of the triplet loss described in~\cite{schroff2015facenet} that incorporates focal attention, see Eq.~\ref{eq:triplet_focal_loss}. The triplet focal loss automatically up-weights hard examples by mapping the euclidean distance to an exponential kernel, penalizing them much more than the easy ones. 
\begin{align}
  \begin{split}\label{eq:triplet_focal_loss}
    \mathcal{L}_{\text{tf}}(\theta) = \displaystyle{\sum_{i=1}^b\max{\left(0, e^{\left(\frac{D_{a_{i},p_{i}}}{\sigma}\right)} - e^{\left(\frac{D_{a_{i},n_{i}}}{\sigma}\right)} + m\right)}},
  \end{split}
\end{align}
\noindent where $\sigma$ is the hyper-parameter that controls the strength of the exponential kernel.

The triplets generation scheme is a critical step that highly impacts the final performance. Traditional methods run their sample strategy over the training set in an off-line fashion, and they do not adapt once the learning process starts. Alternatively, we use an approach for triplets selection based on a semi-hard batch negative mining process, where triplets examples are updated during the training process in each mini-batch, avoiding models to collapse.

The goal of the implemented semi-hard batch negative mining (based on~\cite{parkhi2015deep}) is to choose a negative sample that is fairly hard within a batch but not necessarily the hardest nor the easiest one. For each training step, we select a large set of samples of each class using the current weights of the network. Next, we compute the distances between all positive pairs within this population, \ie $D_{a,p}$, and, for each positive pair, we compute the distance between the corresponding anchor $f(x_a)$ and all possible negative samples $f(x_n)$. Finally, we randomly pick a negative sample that satisfies the following margin criteria, $D_{a,p} - D_{a,n} < m$, to build the final tuples that are used for training at each step, in the so called mini-batch. This mining strategy has two important benefits: 1) we ensure that all the samples included in a training step are relevant for the learning process; and 2) we improve training convergence thanks to the random selection over the negative samples.


In real face anti-spoofing, attackers are constantly engineering new ways to cheat PAD systems with new attacks, materials, devices, \etc. Thus, a classification-like approach is prone to over-fitting to the seen classes and will not generalize well. On the contrary, we follow an anomaly detection based strategy. First, we do not consider the identity of the users as different classes. We define two categories in an anomaly detection setting: 1) the \emph{closed-set}, referring to the classes that can be correctly modeled during training; and 2) the \emph{open-set}, referring to all the classes that cannot be fully modeled by the training set. In face GPAD, genuine samples belong to the closed-set category, while impostors belong to the open-set class, motivated by the scarcity or even the lack of training samples to model certain types of attacks. To achieve this, we fix during training the anchor-positive pair to always belong to the genuine class (\ie the closed-set category) while selecting negative samples from any type of attack (\ie the open-set category) regardless their identity. 

\subsection{Triplet Loss Regularization for a Metric-Softmax}

Recent work~\cite{he2018triplet} demonstrates that the triplet loss, acting as a regularizer of the softmax function, achieves more discriminative and robust embeddings. In our deep anomaly detection based model, we do not focus on the classification task, but instead we aim at obtaining highly representative embeddings to distinguish genuine samples amongst an open-set class of attacks. We thus propose to add the triplet focal loss as a regularizer of a novel softmax function adapted to metric learning, see Eq.~\ref{eq:total_loss}. The proposed softmax formulation, coined as \emph{metric-softmax} ($\mathcal{L}_{\textit{metric\_soft}}$ in Eq.~\ref{eq:loss_metricsoft}), accumulates the probability distribution of each pair within a triplet to be highly separated in an euclidean space. We thus prevent from guiding the learning process towards a binary classification and thus avoiding the well known generalization issues.  
 \begin{align}
  \begin{split}\label{eq:total_loss}
    \mathcal{L}_{\text{anomaly}} = \mathcal{L}_{\textit{metric\_soft}} + \lambda~\mathcal{L}_{\textit{tf}},
  \end{split}\\
  \begin{split}\label{eq:loss_metricsoft}
    \mathcal{L}_{\text{metric\_soft}} = - \displaystyle \sum_{i=1}^{b} \log{\frac{e^{D_{a_{i},p_{i}}}}{e^{D_{a_{i},p_{i}}}+e^{D_{a_{i},n_{i}}}}},
  \end{split}
 \end{align}
 where $\lambda$ is the hyper-parameter to control the trade-off between the triplet focal loss and the softmax loss.
 
The metric learning model proposed obtains a discriminative embedding for every input image. However, we need to provide a posterior probability of whether the image belongs to a genuine sample or to an impersonation attempt. In the experiments, we simply propose to train an SVM classifier with a Radial Basis Function to learn the boundaries between both classes in the feature space.

\subsection{Few-shot a Posteriori Probability Estimation} 
\label{subsec:few-shot}

Often, the inherent dynamic nature of spoofing attacks and the difficulty to access data requires to adapt rapidly to new environments where few samples are available. To deal with this problem, we propose a \emph{few-shot a posteriori estimation} procedure, that does not need any classifier to train on the learned features for decision making in metric learning.

Technically, we proceed to compute the probability of being genuine (see Eq.~\ref{eq:fewshot_da}) as the accumulated posterior probability of the input sample ($x_t$) given two reference sets in the target domain, corresponding to the genuine class $\mathcal{G}$ and the attacks $\mathcal{H}$, respectively. 
 \begin{equation}
   P(x_t~|~\{\mathcal{G}, \mathcal{H}\}) = \displaystyle \sum_{i=1}^{M} \frac{e^{D_{t,g_i}}}{e^{D_{t,g_i}}+e^{D_{t,h_i}}},
   \label{eq:fewshot_da}
 \end{equation}
\noindent where $M$ is the total number of pairs in both reference sets for every attack and for each dataset involved, $t$ sub-index refers to the test image and $\{g_i, h_i\}$ sub-indexes refer to each of the reference samples in the genuine and attack sets, respectively. In order to satisfy the few-shot constraints we choose $M$ to be small in our experiments.

\section{Experimental Results}
\label{sec:experiments}

In this section we present the experiments where our novel approach is compared against three state-of-the-art methods from the literature. The approach in \cite{anomaly_idiap_2018} computes hand-crafted features based on \emph{quality} evidences. They obtain a 139-length feature vector from the concatenation of the quality measurements proposed in \cite{iqa_galbally_2015} and \cite{db_msu_mfsd_2015}. For the second method, we choose ~\cite{color_based_facepad_boulkenafet_2015}, which consists in computing a color-based feature vector of high dimensionality (19998-length) by concatenating texture features based on Local Binary Patterns (LBPs) in two different color spaces (\ie YCbCr and HSV). Finally, the third method is the one proposed in \cite{liu2018learning}, which introduces a two-branch deep neural network that incorporates pixel-wise auxiliary supervision constrained by the depth reconstruction for all genuine samples (attacks are forced to belong to a plane) and the estimation of a remote PhotoPlethysmoGraphy (rPPG) signal to add temporal information. Despite being the state of the art for face anti-spoofing, this model requires to pre-process genuine samples in order to compute the depth estimation and the corresponding rPPG signal, that impacts in the usability and bounds the performance to the methods for depth reconstruction and rPPG estimation. The code for the first two algorithms is based on the reproducible material provided by the authors\footnote{\url{https://github.com/zboulkenafet/Face-anti-spoofing-based-on-color-texture-analysis}} \footnote{\url{https://gitlab.idiap.ch/bob/bob.pad.face/}}. Results for \cite{liu2018learning} are obtained using our own re-implementation of their approach.

\subsection{GRAD-GPAD Framework}

Regardless almost every paper comes with its own reduced dataset~\cite{db_smad_2017, liu2018learning, db_casia_surf_2018, db_rose_youtu_2018}, there is \emph{no agreed upon a PAD benchmark}, and as a consequence, the generalization properties of the models are not properly evaluated. During a brief inspection of the capture settings of available face PAD datasets, one can easily observe that there is no unified criteria in the goals of each of them, leading to a manifest built-in bias. This specificity in the domain covered by most of the datasets can be observed in different scenarios: i) some of them focus on a single type of attacks (\eg masks - 3DMAD~\cite{db_3dmad_2013}, HKBU~\cite{db_hkbu_2016}, CSMAD~\cite{db_csmad_2018}); ii) others focus on the study of different image sources (depth/NIR/thermal) such as CASIA-SURF~\cite{db_casia_surf_2018} or CSMAD; iii) others attempt to simulate a certain scenario like a mobile device setting, where the user hold the device (\eg Replay-Mobile~\cite{db_replay_mobile_2016}, OULU-NPU~\cite{db_oulu_npu_2017}), or a webcam setting, where the user is placed in front of a fixed camera (\eg Replay-Attack~\cite{db_replay_attack_2012}, SiW~\cite{liu2018learning}), or even a stand-up scenario where users are recorded further from the camera (\eg UVAD~\cite{UVAD}).

For our experiments, we propose to use the recently published GRAD-GPAD framework \cite{grad_gpad_gradiant_2019} that mitigates the aforementioned limitations. GRAD-GPAD is the largest aggregated dataset that unifies more than 10 datasets with a common categorization in two levels, to represent four key aspects in anti-spoofing: attacks, lightning, capture devices and resolution. It allows not only a fair evaluation of the generalization properties, but also a better representativity of the face-PAD problem thanks to the increased volume of data. For the sake of the extension of the paper we focus on the evaluation based on the instruments used to perform attacks (\ie PAI - Presentation Attack Instruments) using the categorization in Table~\ref{table:gpad_categorization} (\ie the \emph{Grandtest} protocol).

\begin{table}[htbp]
\footnotesize
\centering
\begin{tabular}{llll} 
  \hline
  \multicolumn{1}{|l|}{Category}                                                      & \multicolumn{1}{l|}{Types}                                                & \multicolumn{1}{l|}{Sub-type}                       & \multicolumn{1}{l|}{Criteria}                                                  \\ \hline
\multicolumn{1}{|l|}{\multirow{9}{*}{\begin{tabular}{l} Presentation\\Attack\\Instrument \end{tabular}}}                    & \multicolumn{1}{l|}{\multirow{3}{*}{print}}          & \multicolumn{1}{l|}{low}       & \multicolumn{1}{l|}{dpi $\leq$ 600pix}                       \\ \cline{3-4} 
\multicolumn{1}{|l|}{}                                        & \multicolumn{1}{l|}{}                                & \multicolumn{1}{l|}{medium}    & \multicolumn{1}{l|}{600 $\textless$ dpi $\leq$ 1000pix}        \\ \cline{3-4} 
\multicolumn{1}{|l|}{}                                        & \multicolumn{1}{l|}{}                                & \multicolumn{1}{l|}{high}      & \multicolumn{1}{l|}{dpi $\textgreater$ 1000pix}              \\ \cline{2-4} 
\multicolumn{1}{|l|}{}                                        & \multicolumn{1}{l|}{\multirow{3}{*}{replay}}         & \multicolumn{1}{l|}{low}       & \multicolumn{1}{l|}{res $\leq$ 480pix}                      \\ \cline{3-4} 
\multicolumn{1}{|l|}{}                                        & \multicolumn{1}{l|}{}                                & \multicolumn{1}{l|}{medium}    & \multicolumn{1}{l|}{480 $\textless$ res $\textless$ 1080pix} \\ \cline{3-4} 
\multicolumn{1}{|l|}{}                                        & \multicolumn{1}{l|}{}                                & \multicolumn{1}{l|}{high}      & \multicolumn{1}{l|}{res $\geq$ 1080pix}                     \\ \cline{2-4} 
\multicolumn{1}{|l|}{}                                        & \multicolumn{1}{l|}{\multirow{3}{*}{mask}}           & \multicolumn{1}{l|}{paper}     & \multicolumn{1}{l|}{paper masks}                          \\ \cline{3-4} 
\multicolumn{1}{|l|}{}                                        & \multicolumn{1}{l|}{}                                & \multicolumn{1}{l|}{rigid}     & \multicolumn{1}{l|}{non-flexible, plaster}            \\ \cline{3-4} 
  \multicolumn{1}{|l|}{}                                        & \multicolumn{1}{l|}{}                                & \multicolumn{1}{l|}{silicone}  & \multicolumn{1}{l|}{silicone masks}                      \\ \hline
\end{tabular}
\caption{Two-tier common PAI categorization in GRAD-GPAD.}
\label{table:gpad_categorization}
\end{table}

\begin{figure*}[htp!]
  \begin{tabular}{>{\centering}m{0.45\textwidth}>{\centering}m{0.45\textwidth}}
    \includegraphics[width=7.6cm]{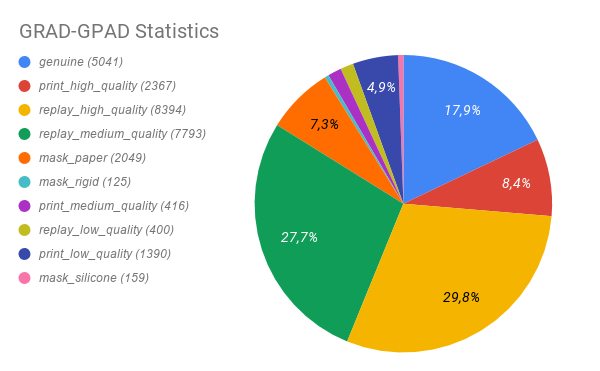}
    &
    \includegraphics[width=8.4cm]{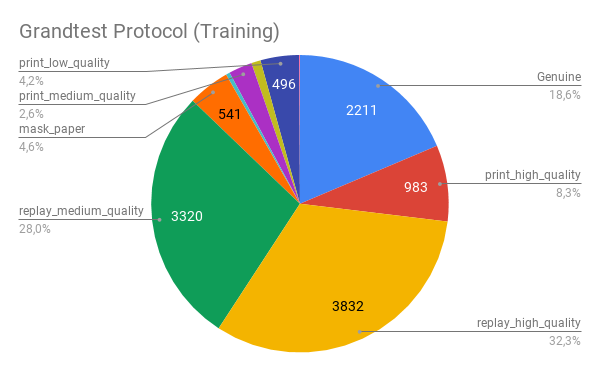}
  \end{tabular}
  \caption{Population statistics for the whole dataset provided in GRAD-GPAD (left) and the training samples statistics for the \emph{Grandtest} protocol (right).}
  \label{figure:GRAD_GPAD_statistics}
\end{figure*}

We conduct all the experiments using the GRAD-GPAD framework, where we add the UVAD dataset \cite{UVAD} to further increase the total number of samples in more than 10k images. In Fig.~\ref{figure:GRAD_GPAD_statistics} we show the population statistics of the whole GRAD-GPAD dataset (left figure) and the training split of the \emph{Grandtest} protocol (right figure).

\subsection{Experimental Setup}

\paragraph{Network Architecture} We use as our backbone architecture a modified version of the ResNet-50 \cite{he2016deep}. 
We stack both RGB and HSV color spaces in the input volume, and feature dimension is fixed to 512. We use \emph{Stochastic Gradient Descent} with \emph{Momentum} optimizer. We start training with a learning rate of $0.01$ using a maximum of 100 epochs. Batch size is fixed to be 12 \emph{triplets}, \ie 36 images per batch. As suggested in the original works, $\sigma$ and $m$ values in Eq.~\ref{eq:triplet_focal_loss} are set to 0.3 and 0.2, respectively. 

\paragraph{Pre-processing} Since our approach follows a frame-based procedure, instead of using the full videos we only pick the central frame of each video. We use as inputs of the network the cropped faces detected using the method proposed in \cite{mtcnn_zhang_2016}.

\paragraph{Metrics} To compare our method with prior works we use the metrics that have been recently standardized in the \textit{ISO/IEC 30107-3}\footnote{https://www.iso.org/standard/67381.html}: \ie False Acceptance Rate (FAR), False Rejection Rate (FRR), Half Total Error Rate ($\text{HTER}=\frac{\text{FAR}+\text{FRR}}{2}$), Attack Presentation Classification Error Rate (APCER), Bonafide Presentation Classification Error Rate (BPCER) and Average Classification Error Rate (ACER). We would like to highlight the importance of the ACER metric because it entails the most challenging scenario, where performance is computed for every attack independently, but it only considers the results for the worst scenario. Thus it penalizes approaches performing well on certain types of attacks. HTER reflects the overall performance of the algorithm in a balanced setting where FAR is equal to FRR, \ie for Equal Error Rate (EER).

\paragraph{Protocols} We evaluate our method on two settings within the GRAD-GPAD framework: 1) intra-dataset; and 2) inter-dataset. For the intra-dataset setting we use the \textit{Grandtest} protocol and for the inter-dataset evaluation we use the leave-one-dataset-out protocols, provided by the framework: the \emph{Cross-Dataset-Test-On-CASIA-FASD} and the \emph{Cross-Dataset-Test-On-ReplayAttack}. In these protocols, one of the datasets (CASIA-FASD and Replay-Attack, respectively) is excluded during training. Results are provided by evaluating the models in the excluded dataset (Test split).

\subsection{Ablation study}
\label{subsec:ablation_study}

The scientific contribution of our work is twofold. First, we introduce a reformulation of the face PAD problem from a deep anomaly detection perspective using metric learning. Second, we propose to use a triplet focal loss as a regularization for a novel softmax loss function adapted to metric learning, coined as ``metric-softmax''. To show the influence of each of these contributions, we conduct the following experiments. We start from a classification-like triplet loss based model, \ie without the anomaly detection approach. This first approach is named as \emph{Baseline} in Table~\ref{table:ablation_study_results}, where tuples for the triplets are selected randomly from the set of classes (genuine + 9 different attacks in GRAD-GPAD). We then incrementally incorporate our contributions. \emph{Model 1} includes the anomaly approach using the triplet loss. In \emph{Model 2} we included the focal attention into the triplet loss formulation. And finally, \emph{Ours} represents the whole pipeline of our system, where the proposed metric-softmax term is added. The results reported in Table~\ref{table:ablation_study_results} show the influence of each contribution in the final performance.

\begin{figure*}[htp!]
  \begin{tabular}{>{\centering}m{0.45\textwidth}>{\centering}m{0.45\textwidth}}
    \includegraphics[width=7.0cm]{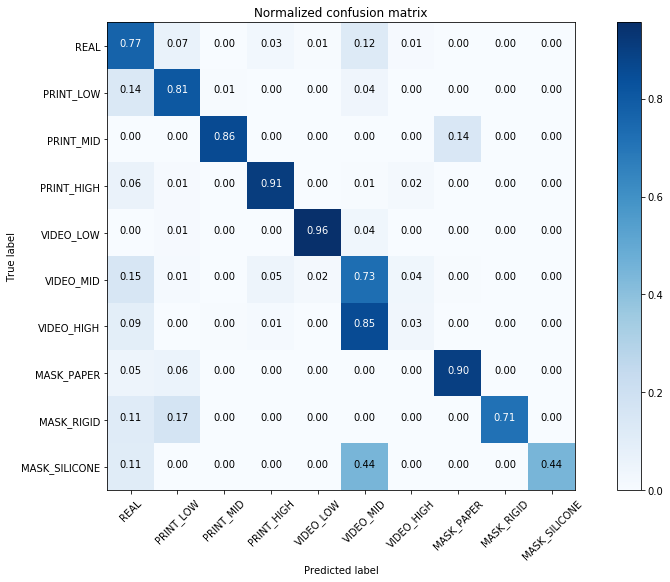}
    &
      \includegraphics[width=7.0cm]{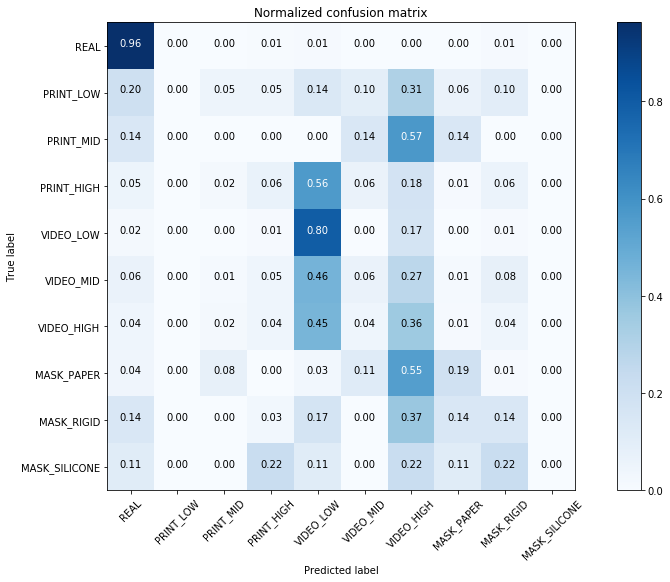}
  \end{tabular}
  \caption{True Positive Rate confusion matrices for the baseline (left) and our approach (right).}
  \label{figure:confusion_matrix}
\end{figure*}

Note that for this ablation study, we use the development split of the \emph{Grandtest} protocol of GRAD-GPAD, and the performance is shown in terms: FAR, FRR and Average Error Rate (AER). Besides, performance is computed using the accumulated metric-softmax distribution described in Eq.~\ref{eq:fewshot_da} with $M=3$ and 
by randomly choosing samples from the training set. 

\begin{table}[htbp]
\small
\centering
\begin{tabular}{|m{1.3cm}|r|r|r|r|}
\hline
\textbf{Algorithm}  & \textbf{AER} & \textbf{FAR} & \textbf{FRR} & $\mathbf{\Delta}$\textbf{AER}  \\ \hline
Baseline    & 17.02 \% & 10.72 \% & 23.33 \% &   -         \\ \hline
Model 1     & 12.46 \% & 24.43 \% & \textbf{0.49 \%} & \textbf{26.8 \%}  \\ \hline
Model 2     & 9.80 \% & 14.87 \%  & 4.74 \% &  \textbf{42.42 \%} \\ \hline
\textbf{Ours}     & \textbf{5.07 \%}    & \textbf{6.38 \%} & 3.77 \%  & \textbf{70.21 \%}\\ \hline
\end{tabular}
\caption{Performance evaluation in the development set of GRAD-GPAD for the different models involved in the ablation study. We also show the relative improvement $\Delta$AER with respect to the baseline.}
\label{table:ablation_study_results}
\end{table}

We show in Table~\ref{table:ablation_study_results} that, when we incorporate the focal attention into the triplet, \ie \emph{Model 2}, we achieve a relative improvement of 42.42\% in terms of AER. This aspect reveals the importance of a mining strategy in the learning process. Finally, the introduction of the proposed metric-softmax term, achieves a remarkable 
relative improvement of~70.21\% of AER.

Furthermore, we show in Fig.~\ref{figure:confusion_matrix} the True Positive Rate (TPR) confusion matrices for the \emph{Baseline} (left) and our approach (right). We assess that, with the anomaly detection approach, we are able to highly differentiate genuine from impostor samples, regardless the classification of the attack instrument. Note that the \emph{baseline} obtains poor performance for genuine samples classification, despite classifying correctly the different attacks, which highly penalizes its global performance.

\subsection{Intra-Dataset Evaluation}

In order to fairly compare our approach with the state-of-the-art methods, we train an SVM-RBF classifier for each of them using their corresponding features. Additionally, for the \emph{Auxiliary} model \cite{liu2018learning}, we report the results just using the L2-Norm from the depth map (\emph{Auxiliary$^*$}), as it is proposed by the authors in their original work. For all the experiments, we use $M=3$ in Eq.~\ref{eq:fewshot_da} for the few-shot a posteriori probability estimation (\emph{Ours$^\dag$}) experiment. 
Note that, both the original method proposed in~\cite{liu2018learning} (\emph{Auxiliary$^*$}) and our approach with a posteriori estimation, do not need to use any classifier with the learned features.

\begin{table}[hptb!]
  \small
  \centering
  \begin{tabular}{|l|r|r|r|r|}
    \hline
    \textbf{Algorithm}  & \textbf{HTER}    & \textbf{ACER}     & \textbf{APCER}    & \textbf{BPCER}   \\ \hline
    Quality \cite{anomaly_idiap_2018}     & 23.21 \%  & 36.96 \%          & 50.51 \%          & 23.42 \%         \\ \hline
    Color \cite{color_based_facepad_boulkenafet_2015}      & 7.87 \%   & 19.21 \%          & 28.57 \%          & 9.84 \%          \\ \hline
    Auxiliary \cite{liu2018learning}  & 5.92 \%   & 37.89 \%          & 66.67 \%          & 8.55 \%          \\ \hline
    Auxiliary$^*$ & 6.52 \%   & 31.81 \%          & 53.33 \%          & 10.44 \%         \\ \hline
    Ours        & \textbf{5.41 \%} & \textbf{10.14 \%} & 14.29 \% & \textbf{5.99 \%} \\ \hline
    Ours$^\dag$  & 5.45 \%  & 10.42 \%          & \textbf{14.28} \%          & 6.55\%          \\ \hline 
  \end{tabular}
  \caption{Intra-dataset results on the \emph{Grandtest} protocol.}
  \label{table:intradb_results}
\end{table}

Results in Table~\ref{table:intradb_results} demonstrate that both our novel approaches outperform the state-of-art methods, even using the most challenging metric (ACER). These results highlight that the learned feature space 
has a high discrimination capability and that our model performs the best. 

\subsection{Inter-Dataset Evaluation}

In order to assess the generalization capabilities, we perform two cross-dataset evaluations where a whole dataset is excluded from the training step. In the first experiment, we leave out CASIA-FASD \cite{db_casia_fasd_2012} for the test set. In the second one, ReplayAttack~\cite{db_replay_attack_2012} is excluded during learning. In both experiments, none of the samples from the test dataset are used neither in the training set nor in the development set.

\subsubsection{Test on CASIA-FASD}

\begin{figure}[htbp]
  \includegraphics[width=0.44\textwidth]{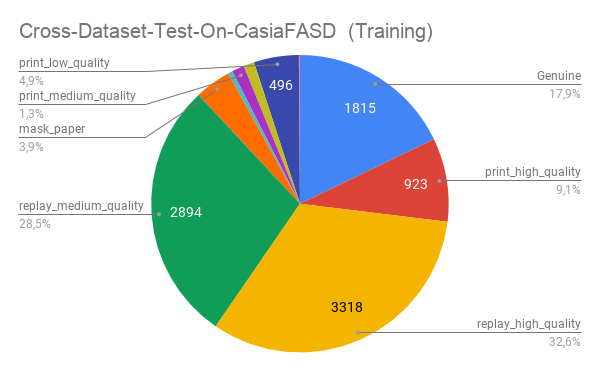}
  \caption{Training samples statistics for the \emph{Cross-Dataset-Test-On-CASIA-FASD} protocol.}
  \label{figure:casia_test_classes}
\end{figure}

As it is shown in Fig.~\ref{figure:casia_test_classes}, the training set for this experiment includes all types of attacks, however the domain is different (\ie different environments, lighting conditions, capture devices, \etc). CASIA-FASD is one of the smallest datasets for face anti-spoofing samples. Therefore, considering only its test set for the evaluation would highly penalize the performance in case of miss-classification. This fact is reflected in Table~\ref{table:interdb_casiatest_results}, where performance significantly drops in all methods, except for our approach, where we are able to keep a reasonable good performance: from an ACER of 10.14\% (see Table~\ref{table:intradb_results}) to 16.8\%. 

\begin{table}[hptb!]
  \small
  \centering
  \begin{tabular}{|l|r|r|r|r|}
    \hline
    \textbf{Algorithm} & \textbf{HTER}  & \textbf{ACER} & \textbf{APCER} & \textbf{BPCER} \\ \hline
    Quality \cite{anomaly_idiap_2018} & 40.90 \%   & 47.38 \%     & 65.56 \%   & 29.21 \%  \\ \hline
    Color \cite{color_based_facepad_boulkenafet_2015} & 22.17 \%  & 25.69 \%  & 26.67 \%   & 24.72 \%    \\ \hline
    Auxiliary \cite{liu2018learning}   & 28.60 \%  & 29.71 \% & 12.22 \%  & 47.19 \% \\ \hline
    Auxiliary$^*$  & 25.42 \%  & 26.90 \% & 12.22 \%  & 41.57 \% \\ \hline 
    Ours  & \textbf{16.74 \%} & \textbf{16.80 \%} & \textbf{10.00 \%} & \textbf{23.60 \%} \\ \hline
    Ours$^\dag$  & 17.56 \% & 18.48 \% & \textbf{10.00} \% & 26.97 \% \\ \hline
  \end{tabular}
  \caption{Inter-dataset results evaluated on CASIA-FASD.} 
  \label{table:interdb_casiatest_results}
\end{table}

In Table~\ref{table:interdb_casiatest_results} we show that HTER and ACER values for our approach are almost the same. We argue that, despite the domain shift introduced by this protocol, the learned embeddings during training are robust enough to generalize in this setting. Instead, the other methods in the experiment are highly penalized, showing that they tend to overfit on the training set to a greater extent.

Besides, we show that our few-shot a posteriori estimation pipeline (Ours$^\dag$) achieves similar performance compared to the SVM version in this Test on CASIA-FASD setup. Thus, we assess that the learnt embedding space generalizes enough so that we can avoid using a classifier with the feature vectors and estimate the a posteriori probability by simply using $M=3$. This classiffier-free model is also able to outperform all state-of-the-art methods, including \emph{Auxiliary$^*$} that neither requires a classifier.


\subsubsection{Test on Replay-Attack}

\begin{figure}[htbp]
  \includegraphics[width=0.44\textwidth]{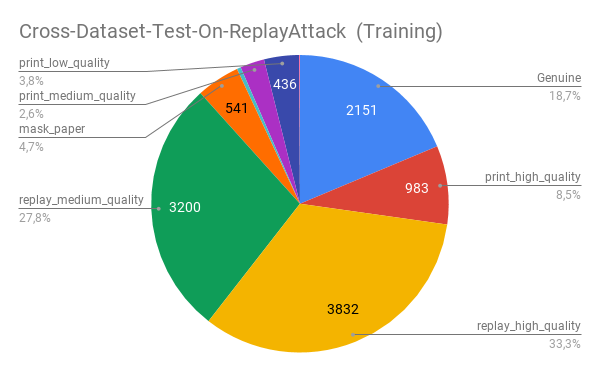}
  \caption{Training samples statistics for the \emph{Cross-Dataset-Test-On-ReplayAttack} protocol.}
  \label{figure:replayattack_test_classes}
\end{figure}

The motivation behind selecting to leave out the Replay-Attack dataset is to show the impact in the performance of face-PAD algorithms of unseen attacks belonging to a new domain: this dataset contains all the samples for \emph{replay-low-quality} attacks (see Fig.~\ref{figure:replayattack_test_classes}). This entails a far more challenging scenario.

\begin{table}[hptb!]
\small
\centering
\begin{tabular}{|l|r|r|r|r|}
  \hline
  \textbf{Algorithm} & \textbf{HTER}     & \textbf{ACER}     & \textbf{APCER}    & \textbf{BPCER}    \\ \hline
  Quality \cite{anomaly_idiap_2018}  & 37.35 \%          & 47.02 \%     & \textbf{42.14 \%} & 51.90 \%          \\ \hline
  Color \cite{color_based_facepad_boulkenafet_2015}     & 34.51 \%          & \textbf{43.35 \%}     & 51.25 \%          & 35.44 \%          \\ \hline
  Auxiliary \cite{liu2018learning}  & 35.62 \%          & 45.62 \%     & 68.75 \%          & 22.50 \%          \\ \hline
  Auxiliary$^*$ & 37.87\%           & 47.50 \%     & 72.50 \%          & 22.50 \%          \\ \hline
  Ours       & \textbf{25.00} \%          & 45.62 \%     & 71.25 \%          & \textbf{20.00} \% \\ \hline
  Ours$^\dag$   & 25.25 \%          & 45.62 \%          & 71.25 \%          & \textbf{20.00 \%} \\ \hline    
\end{tabular}
\caption{Inter-dataset results evaluated on Replay Attack.}
\label{table:interdb_replayattacktest_results}
\end{table}

The results reported in Table~\ref{table:interdb_replayattacktest_results} show a severe drop of performance for all the methods, specially for ACER, where all the approaches are highly penalized by the unseen attack and achieves performance close to random choice. This fact is due to the addition of a new attack that has never seen before in combination with a strong domain change, highly impacting on APCER (\ie the attack classification). Interestingly, our proposal based on few-shot a posteriori 
estimation keeps exactly the same performance compared with our method with an SVM, again assessing that we can replace the classifier using a few samples. Besides, we obtain the best overall performance HTER and the best BPCER (ACER is close to random choice for all the methods).




\section{Conclusions}
\label{sec:conclusions}

In this work we introduce a novel approach that addresses the problem of generalization in face-PAD, following an anomaly detection pipeline. We leverage deep metric learning to propose a new ``metric-softmax'' loss that applied in conjunction with the triplet focal loss drives to more robust and generalized features representations to distinguish between original and attack samples. We also propose a new a posteriori probability estimation 
that prevents us from the need of training any classifier for decision making. 
With a thorough experimental evaluation in the challenging GRAD-GPAD framework we show that the proposed solution outperforms prior works by a considerable margin.

\paragraph{Acknowledgements} We thank our colleagues of the Biometrics Team at Gradiant for their valuable contributions. Special mention to Esteban Vazquez-Fernandez, Juli\'an Lamoso-N\'u\~nez and Miguel Lorenzo-Montoto.

{\small
\bibliographystyle{ieee_fullname}
\bibliography{egbib}
}


\end{document}